# Graph representation learning on street networks

Mateo Neira[1,2] and Roberto Murcio[1,3]

[1]Centre for Advanced Spatial Analysis, University College London, London, United Kingdom; [2]The Alan Turing Institute, London, United Kingdom; [3]Department of Geography, Birkbeck, London, United Kingdom;

## ABSTRACT

Streets networks provide an invaluable source of information about the different temporal and spatial patterns emerging in our cities. These streets are often represented as graphs where intersections are modelled as nodes and streets as links between them. Previous work has shown that raster representations of the original data can be created through a learning algorithm on low-dimensional representations of the street networks. In contrast, models that capture high-level urban network metrics can be trained through convolutional neural networks. However, the detailed topological data is lost through the rasterisation of the street network. The models need help to recover this information from the image alone, failing to capture complex street network features. This paper proposes a model capable of inferring good representations directly from the street network. Specifically, we use a variational autoencoder with graph convolutional layers and a decoder that outputs a probabilistic fully-connected graph to learn latent representations that encode both local network structure and the spatial distribution of nodes. We train the model on thousands of street network segments and use the learnt representations to generate synthetic street configurations. Finally, we proposed a possible application to classify the urban morphology of different network segments by investigating their common characteristics in the learnt space.

Keywords:   Street networks; representational learning; graph convolutional networks

## INTRODUCTION

Street networks are a critical component of the urban fabric, and they have a profound impact on the way that cities function. They are the arteries through which people and goods move, and they play a pivotal role in shaping cities' social, economic, and physical landscape. Understanding how street networks evolve and their use can help us better comprehend urbanization's complex processes. Given the importance of cities, urbanization, and their role in tackling environmental, economic, and social challenges, street networks have become an object of scientific study over the last fifty years across various disciplines, including transport and urban planning, geography, and physics. Over this time, there has been a wealth of work to quantitatively analyse street networks and build models capable of generating networks that exhibit the same empirical features. Many of these studies represent street networks as graphs where street intersections are modelled as vertices, and street segments as edges, Marshall et al. (2018) provide an overview of the different types of street network models found in the literature. This type of representation has allowed researchers to apply methods from network science and complexity science to 1) understand the structural Strano et al. (2012), topological Louf and Barthelemy (2014), hierarchical Arcaute et al. (2016) and fractal Murcio et al. (2015) properties of street networks, and 2) to be able to model their evolution and growth Barthelemy and Flammini (2006, 2008).´

New large-scale open datasets, like those generated through crowdsourced volunteered geographical information OpenStreetMap contributors (2019), and the development of machine learning methods that can extract useful information from these vast amounts of data, research has emerged in the last five years that aims to capture the full breadth and complexity of existing urban structure by complementing traditional methods and generating a better understanding of these systems.

Traditional methods of analysing street networks represented on graphs rely on user-defined heuristics to extract relevant features that can be analysed (e.g. degree statistics or centrality measures). By leveraging large-scale open datasets with machine learning methods, we can automatically learn to encode street network structure into a low-dimensional latent feature vector (also called embeddings). These representation-learning approaches remove the need for feature engineering and have been shown to outperform traditional methods for many applications Bengio et al. (2013); LeCun et al. (2015)

In the case of street networks, convolutional neural networks have been used on images representing street networks to create low-dimensional embeddings. They have been shown to capture relevant information about urban structures. For example, a convolutional autoencoder (CAE) was used in Moosavi (2022) to learn an embedding to cities across the world that was used to cluster cities based on their urban structure using a self-organizing map. Similarly, Kempinska and Murcio (2019) showed that a variational autoencoder (VAE) could be used to learn embeddings of different cities and measure similarity across street networks. Using a similar approach Law and Neira (2019) developed a method called ConvPCA to create interpretable latent features that could be interrogated by using a combination of geographical mapping and latent space perturbations. The later work showed that these approaches fail to capture topological features of the street networks and that most of the information captured relates to street network density.

Generative models have also been used to generate synthetic street networks. For example, Variational Autoencoder trained on street network images has been used by sampling from the latent space z Kempinska and Murcio (2019). However, this resolution is low and fails to capture fine-grain detail of local streets. Generative Adversarial Networks Hartmann et al. (2017) have also proposed to generate a multitude of arbitrary-sized street networks that faithfully reproduce the style of the original dataset. Although these models have been shown to capture general patterns of street networks, the resultant latent spaces fail to capture the topological properties of the data. More work needs to be done regarding how these latent spaces relate to established street network measures. Additionally, using these methods for street network generation requires a post-processing step to turn the results images back into a graph representation that can introduce additional errors in the results.

We introduce a model capable of inferring good representations directly from the street network that can generate synthetic street networks. Specifically, we use a model based on a variational autoencoder with graph convolutional layers Kipf and Welling (2016a) to address some of the shortcomings of learning low-dimensional vector representations of street networks that can be used in downstream tasks, such as the classification of urban morphology of different cities, and for street network generation. The model can encode both the local network structure and the spatial distribution of nodes. We train the model on 39,000 towns and cities and use the learnt representations to classify the urban morphology of different places and investigate their relation to established street network metrics such as circuity, average street length, and an average number of edges per node to evaluate the capacity of the model to generate synthetic networks.

## METHODS AND MATERIALS

This work aims to learn the embeddings of street networks for use in various downstream tasks such as street network classification and generation of synthetic street networks. The learnt embeddings should capture the street networks' spatial and topological structure. We retrieve street network data for cities worldwide and create an undirected network $G = (V,E)$ where $V$ represents street junctions and $E$ streets connecting them, along with a node feature matrix $X$ containing the coordinates (latitude and longitude) of each node. Given the adjacency matrix *A* of *G* and the node feature vector *X*, we want to learn a function $f: G \to \mathbb{R}^d$ which projects each street network into a d-dimensional latent space $\mathbb{R}^d$.

We do this by learning a distribution over graphs *G* with node features *X* from which we can generate new examples. We split the modelling task into two parts: 1) generating graph nodes with their coordinate pairs *X* and 2) generating an adjacency matrix *A* for the given nodes. We can formulate this as follows:

$$p(G) = p(A,X)$$

$$= p(A|X)p(X)$$

We use separate node and adjacency models, the first uses an auto-regressive approach, and the second is a variational graph auto-encoder. In this way, we create two embeddings, one for individual nodes that encode its spatial distribution and the second for the entire network, which encodes the spatial distribution of all nodes and the topological network structure. We can generate synthetic street networks by sampling from the node model and then passing the resulting nodes as inputs to the variational graph auto-encoder, as shown in Figure 3.

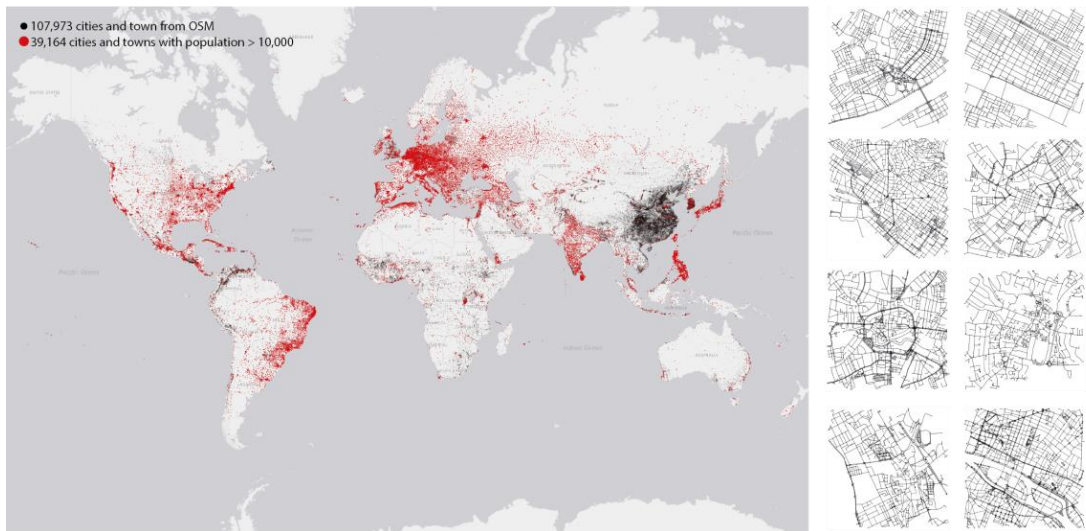

**Figure 1.** The geographical location of cities and towns used for training and testing (left), a sample of randomly selected street networks of selected cities (right).

**Data processing**
The street network dataset is taken from OpenStreetMap (OSM) OpenStreetMap contributors (2019), a publicly available geographic database of the world. Street networks are represented as piecewise-linear polylines with an associated highway label to distinguish them from other geographical structures. The raw road network extracted from OSM is represented as a vector containing their geo-coordinates in the WGS84 projection. We query all the cities and towns for a total of 107,973 by extracting all features with a *place* tag equal to *town* or *city*. These centres do not necessarily correspond to defined cities but are important urban centres with a good range of shops and facilities. We also extract any information related to the population for these centres that we use to subset the initial data for different tasks. As per OSM, 34% of nodes tagged as towns have a population tag, and the median value in the database varies by country, from as low as 300 to as high as 35,000. 95% of towns have a population=* value between 1000 and 70,000. In the case of nodes tagged as cities, 63% of having a population tag, of which the median value is 130,000, and 95% of them have a population value over 20,000. To train the model, we use a subset of cities containing a population tag valued over 1,000. This results in a total of 39,364 cities and towns, the spatial distribution of these cities and towns is shown in Figure 1. Of these networks, the mean number of nodes within the 1sqkm is 430, and the mean number of edges is 1,100.

The distribution of the number of nodes and edges in the entire dataset is shown in Figure 2.

For each of the selected cities and towns, we download the street network within a 1km x 1Km box at the centroid of each place using osmnx Boeing (2017). For each grid, we retrieve a graph G = (V, E) where each vertex v corresponds to a street intersection, and e edge corresponds to a street segment. Each street network is first re-projected from the given spherical coordinates to meters using the UTM projection for the given area. Then, the street networks are further simplified by joining nodes closer to a threshold of 10m.

**0Node model**

The node model aims to estimate a distribution over sequences of nodes. To facilitate this, we impose a node ordering of lowest to highest by the *y*-coordinate. If nodes have the same *y* value, we order these by their *x*-coordinate. After imposing a node ordering, we obtain a flattened coordinate sequence $C^{seq}$ by concatenating the coordinate pairs $(x_i, y_i)_i$. The coordinate sequence can be decomposed as a joint distribution over its elements as the product of a series of conditional coordinate distributions. We model this distribution using a transformer Vaswani et al. (2017) in an auto-regressive manner to facilitate learning the spatial distribution of node coordinates; we centre each street network at (0,0) and normalise both x and y coordinates such that the diagonal of their bounding box is equal to 1. Once the nodes are centred and normalised, we apply a uniform 8-bit quantisation that allows us to model the nodes' coordinate values as categorical distributions.

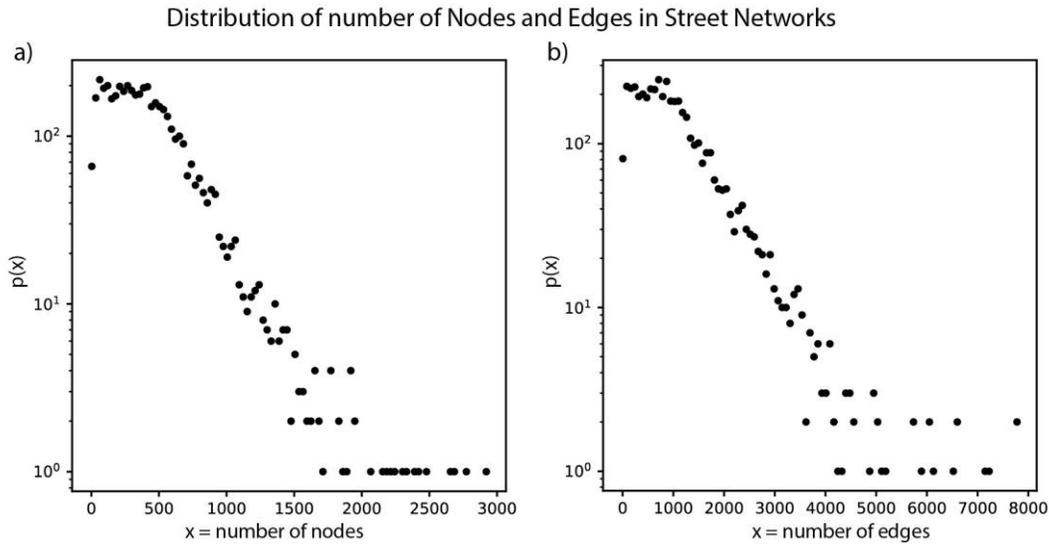

**Figure 2.** Network size distribution of street network data. a) nodes b) edges.

A similar approach has been taken to model 3D meshes Nash et al. (2020) and to discretise continuous signals Van Oord et al. (2016), with the benefit of being able to express arbitrary distributions. We find that the spatial extent of 1km 8-bit quantisation does not reduce the overall size of the network, as all nodes fall into different bins.

Similar to PolyGen Nash et al. (2020), we use three embeddings that are learnt during training for each input token: a coordinate embedding, which indicates where the input token is an x or y coordinate; a positional embedding, which indicates which vertex in the sequence the token belongs to, and a value embeddings, which expresses a quantised coordinate value. The output at the final step of the transformer is the logits of the distribution of the quantised coordinate values. We use the embeddings learnt as feature vectors for each node as inputs to the variational graph auto-encoder model.

**Graph Auto-Encoder model**

Once we have a series of node embeddings, we want to estimate the edge distributions by estimating the adjacency matrix for a given set of nodes. To achieve this, we use a variational graph auto-encoder Kipf and Welling (2016b), a framework for unsupervised learning on graph-structured data based on the variational auto-encoder Kingma and Welling (2013). The model can learn latent representations for undirected graphs using an adjacency matrix and a node feature matrix. We use a series of graph convolutional networks (GCN) Kipf and Welling (2016a) to learn local and global structures present in the data.

Given the undirected unweighted street network graph G = (V, E) with $N = |V|$ nodes, we introduce an adjacency matrix A of $G$ with all the diagonal elements equal to one. Its degree matrix D. We further introduce stochastic latent variables $z_i$, summarized in an $N \times F$ matrix Z. Node features containing the embeddings learnt from the node model are summarized in an $N \times 128$ matrix X. The variational graph auto-encoder takes in the adjacency matrix A and the node embeddings X and learns an embedding that encodes both the topological and node positions.

The encoder of the VGAE is composed of graph convolutional networks (GCNs). The model takes the adjacency matrix A and the node embeddings X as inputs and generates latent variables Z. The first layers of the GCN are defined as:

$$\text{GCN}(A,X) = \tilde{A} \, \text{ReLU}(\tilde{A} X W_0) W_1 \tag{1}$$

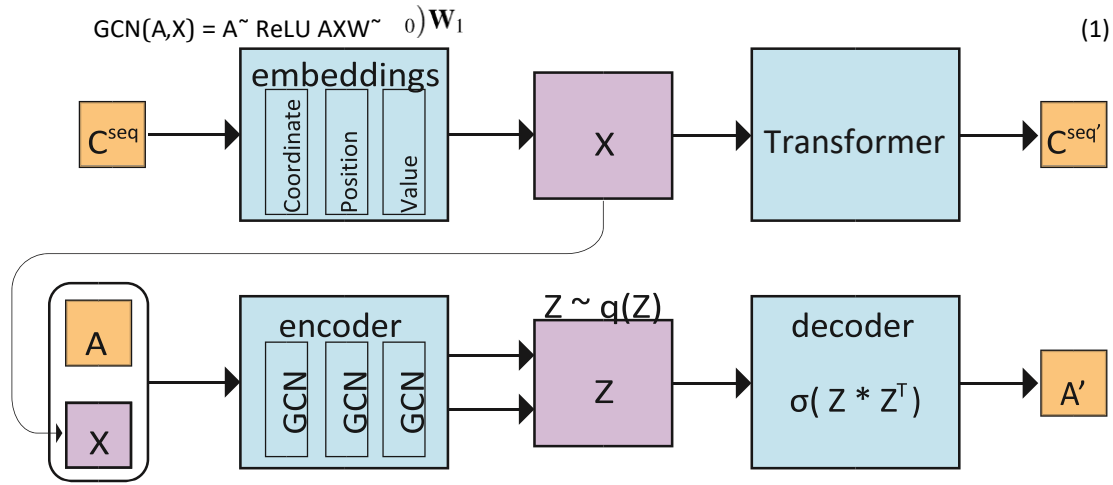

**Figure 3.** Variational Graph Autoencoder takes an adjacency matrix and a node embedding matrix as input condenses both matrices to a lower-dimensional encoding (middle) and finally reconstructs the graph given the encoding (right).

$$\tilde{A} = D^{-\frac{1}{2}} A D^{-\frac{1}{2}} \tag{2}$$

A' is the symmetrically normalised adjacency matrix, and *ReLU* is a rectified linear activation function that has been shown to work well for graph neural networks and is easier to learn than other types of activation functions.

The last layer of the encoder generates $\mu$ and log $\sigma^2$. These are used to calculate $Z$ using a parameterization trick $Z = \mu + \sigma * \varepsilon$, where $\varepsilon \sim N(0,1)$. Here, $\mu = GCN_\mu(A,X)$ is the matrix of mean vectors $\mu_i$; similarly log $\sigma = GCN_\sigma(A,X)$.

The decoder is given by a generative model defined by an inner product between latent variables. The output of the decoder is a reconstructed adjacency matrix $A'$, which is defined as:

$$p(\mathbf{A'}|\mathbf{Z}) = \prod_{i=1}^{N} \prod_{j=1}^{N} p\left(A'_{ij}|\mathbf{z}_i, \mathbf{z}_j\right), \text{ with } \quad p\left(A'_{ij} = 1 | \mathbf{z}_i, \mathbf{z}_j\right) = \sigma(\mathbf{z}_i^\top \mathbf{z}_j), \quad (3)$$

where $A'_{ij}$ are the elements of $A'$ and $\sigma(\cdot)$ is the logistic sigmoid function.

We optimize the variational lower bound $L$ w.r.t. the variational parameters $W_i$:

$$L = \mathbb{E}_{q(\mathbf{Z}|\mathbf{A}),\mathbf{X}}\left[\log p(\mathbf{A}|\mathbf{Z})\right] - KL\left[q(\mathbf{Z}|\mathbf{A},\mathbf{X}) || p(\mathbf{Z})\right], \quad (4)$$

where $KL[q(\cdot)||p(\cdot)]$ is the Kullback-Leibler divergence between $q(\cdot)$ and $p(\cdot)$. We further take a Gaussian prior $p(Z) = \prod_i p(z_i) = \prod_i N(z_i|0,I)$. For very sparse A. We perform full-batch gradient descent and use the *reparameterization trick* Kingma and Welling (2013) for training.

The node and variational auto-encoder models were implemented using PyTorch with PyTorch Geometric library for graph convolutional layers Fey and Lenssen (2019). Both networks are trained using an 80-20 training-testing split of the entire dataset. The final embeddings we obtain for downstream tasks are created by concatenating embeddings from Z and further reducing the size of the resulting latent variables by applying a dimensionality reduction that maximises the variance across the entire training dataset.

 **Results**

We test the performance of our model by evaluating its capabilities to generate synthetic street networks. We empirically evaluate network reconstruction quality by sampling from the learned lower dimensional latent space and recreating synthetic street networks. The recreated street networks are compared against the street networks in our testing data set containing 20% of the data. A sample of synthetic street networks

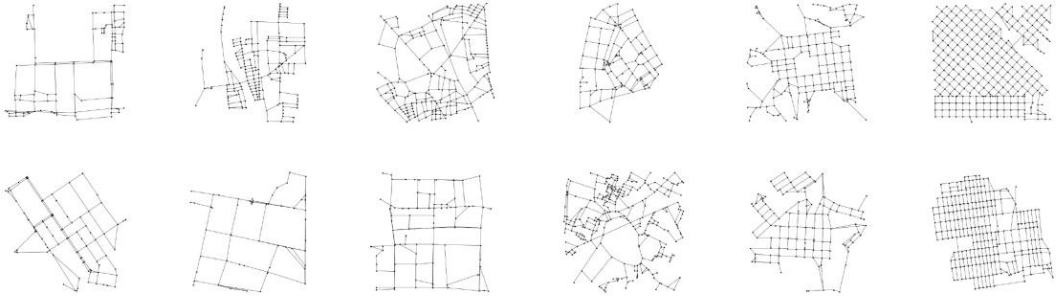

**Figure 4.** Samples of generated street networks are first sampled from the node model to produce a sequence of nodes along with their x and y coordinates and embeddings. The embeddings are inputs to the variational graph auto-encoder model to obtain an adjacency matrix.

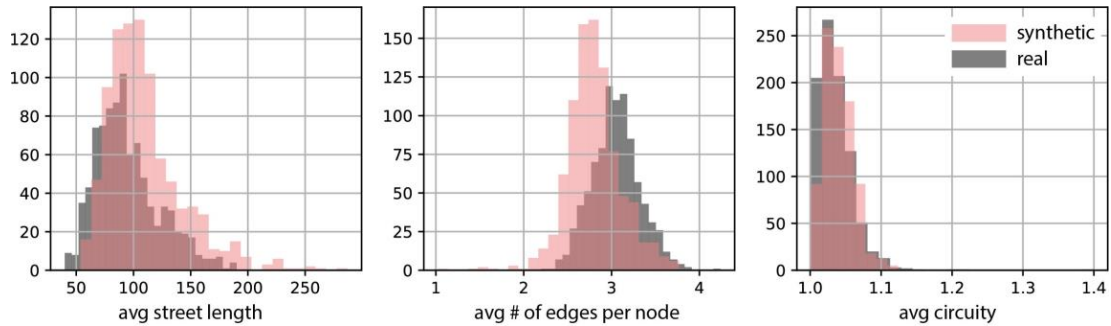

**Figure 5.** Distribution of the topological properties of real and generated samples of street networks. From left: average street length, middle: average number of streets per intersection, and right: average circuity defined by the ratio between the network distance and Euclidean distance between two nodes.

Generated from the model is shown in Figure 4. Our model performs well in generating synthetic street networks and captures both topological and geometric properties of real street networks well. However, in some cases, the synthetic data contains some artefacts, such as small triangulated intersections and a series of degree 2 nodes forming one continuous street, we hypothesise that this may be due to the size of the lower dimensional latent space and training time, and further work is needed to test improvements to the model.

*Topological features*

We compare the distribution of certain graph summaries from samples from our model to those of real street networks. If our model closely matches the true data distribution, we expect these summaries to have similar distributions. We draw 1,000 samples from our model and 1,000 from our testing set. Figure 5 shows the distribution of the average street length, the average number of edges per node, and the average circuity (network distance divided by the Euclidean distance between two neighbouring nodes). Although these are coarse-grained descriptions of street networks, our model has a similar distribution over each street network statistic. We note that the generated samples tend to have a slightly higher average street length and fewer edges per node. The latter is due mainly to the tendency of the model to produce a higher number of degree two nodes.

*Geometric features*

We also compared the distribution of the geometric properties of the generated street networks. For this task, we look at the faces of the planar graph formed by the street - which could be considered city blocks. For each of the graphs of both the testing data and the generated synthetic street networks, we extract the faces of the graph and calculate simple measures to characterise their geometry to compare their distributions. We calculate three measures for all the blocks formed by the street network: average block area in square meters, average form factor, the ratio between the area of the block and the area of its

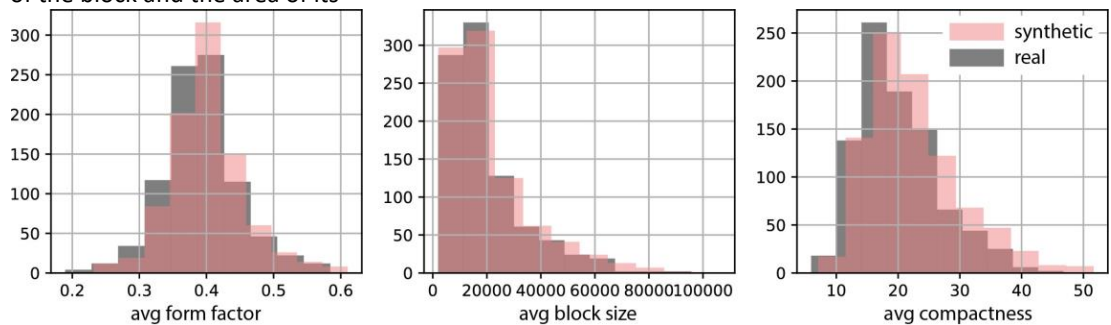

**Figure 6.** Distribution of the geometric properties of real and generated samples of street networks. Left: average form factor defined by the ratio of a block's shape and the area of its circumscribed circle; middle: average block size; and right: average compactness measured by the ratio between the block's perimeter length and its area. For each graph, blocks are defined by the faces of the spatially embedded graph.

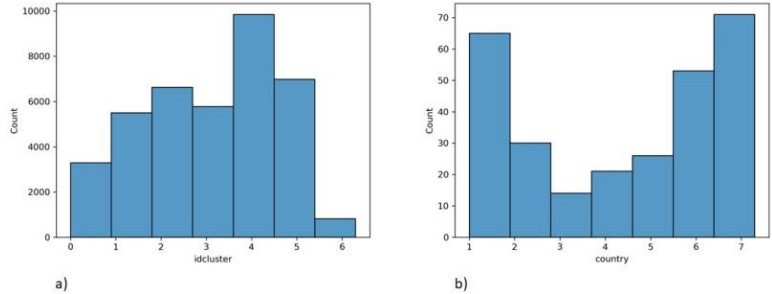

**Figure 7.** a)Cluster membership distribution. Clusters 2 to 5 are distributed evenly across the dataset, while cluster 6 is the less common type. b) Variety of cluster type by country. There is a significant number of countries with only one type of cluster. However, most of these are small countries with only one street network represented in our data.

Circumscribed circle, and the average compactness, are measured as the ratio of each block's perimeter length and area. Results are shown in Figure 6. We see that the distribution of the synthetic data's geometric properties closely matches that of the real networks.

**Empirical study of the learnt embedding - street network classification**
*Clustering*
Following Kempinska and Murcio (2019), we performed a distance and clustering analysis over the latent representations of each road network. Under the premise that the latent space encompasses the main characteristics of the topological structure of the street network, we first cluster all the studied networks using a KMeans approach, with k=7 [1]. Figure 7) shows the distribution of the cluster membership among all the street network sections studied. A quarter of the networks belong to cluster number 4, while only 2% belong to cluster number 6 (Fig 7. a). This distribution provides evidence about how street networks, despite their historical, planning and political differences, can be classified according to their topology and, ultimately, their topological function. A sample of cluster types 4 and 6 is shown in Fig 8

Moving the analysis to individual countries, we observed that various clusters per country tell us a different history. Inside a country, we can find different types of clusters (Fig 7. b). Most countries have six or seven types of clusters, showing the particular functions and genesis of different segments of the street network in a single city. In the case of countries with only one type of cluster, there is an evident bias towards places with only one street network in the data so that we can derive no significant conclusion for this cluster of countries.

Interestingly, for the 235 countries studied, there is a clear tendency to have a "preferred" cluster, i.e., each cluster membership distribution has a unique mode value. As stated, the main point of this work is not the cluster analysis but the methodology to construct a latent representation of street networks; however, Figure 9 (which shows the most common membership by country) produces several results indicating striking similarities between networks among countries. For example, Portugal, Brasil and Mexico share the same local street network structure with some parts of Eastern Europe and Russia. Also, we found that cluster 3 is the preferred cluster for China, the Philippines, India, Turkiye and, surprisingly, Central America. The fact that we can observe a completely different set of clusters for the countries between India and China presents an interesting research venue that should be addressed elsewhere. Finally, we observed that the segments of street networks located in the geographical centres of large cities (Los Angeles, Mexico City, Mumbai, El Cairo, Sao Paulo, Shanghai, for example) group in the less~ common cluster, cluster number 6.

---
[1] The number of clusters was defined using the Elbow method

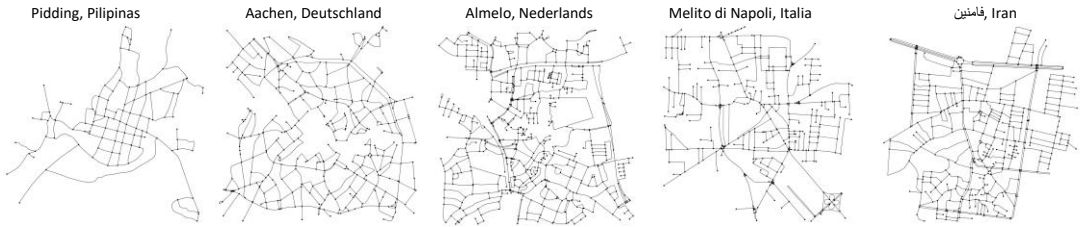

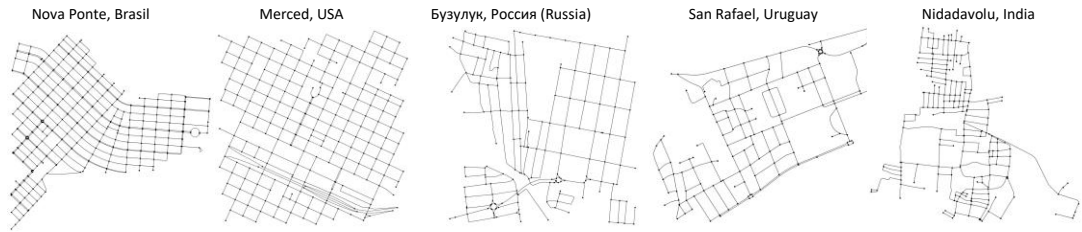

**Figure 8.** A random sample of ten street networks clustered by their membership. Top: Cluster 4 - most common type; Bottom: Cluster 6 - less common. These results suggest that non-regular grid street areas are more common across cities than regular ones and that the latent space is picking up these topologies. Hence, after clustering, they belong to the same membership.

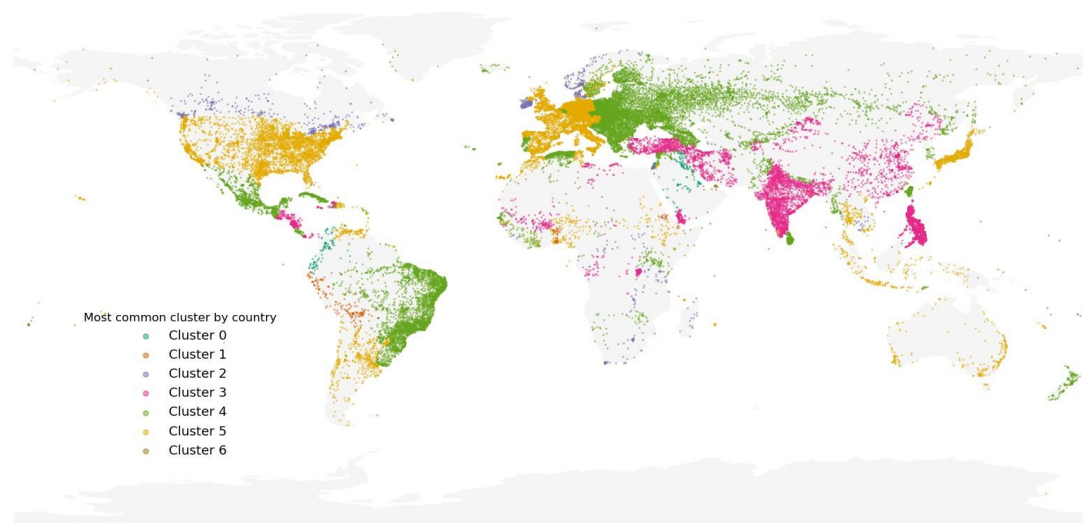

**Figure 9.** Most common cluster by country. Some spatial patterns start to emerge in the cluster distribution mode by country. For example, India, China, Indonesia and Turkiye share the same structure.

**City's street orientation by cluster membership**

To further evaluate the relevance of our clusters, we look into the street's orientation order, following. Boeing (2018). Our results are different from Boeing (2018) as the analysis was made over the whole city, not over smaller street segments as in our case. Nevertheless, we found that cities from the same cluster tend to hold the same cardinal direction. For example, the cluster membership mode in Canada is 2, and 90% of all cluster 2 street networks in Canada have a North-East-South-West orientation, like the example in Fig 10. On the other hand, American cities belonging to cluster 5 have an orthogonal orientation, while two small Mexican networks have a strong North-East orientation. As before, these findings need further validation out of the scope of this paper.

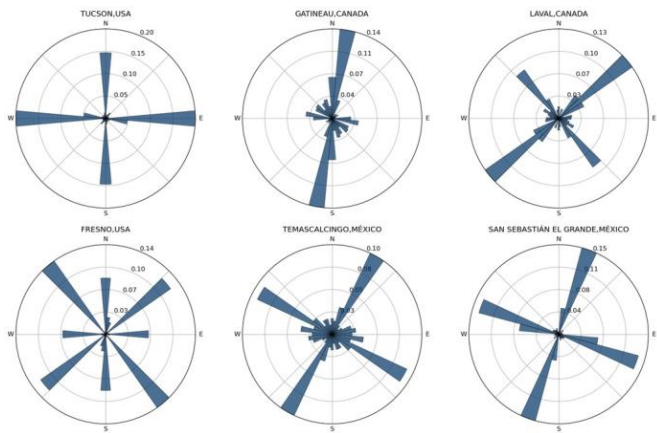

**Figure 10.** Six cities' orientations examples for North American street networks. Tucson and Fresno depict the strong orthogonal angular structure found in many American cities; the two Canadian and Mexican cities present a similar orientation (N-E-S-W) and structure, although the two Mexican are indeed similar in terms of the angles.

## CONCLUSIONS

In this work, we present a Graph Variational Auto-Encoder (GVAE) model to learn latent representations of street networks. The model can infer good latent representations that encode local network structure and the spatial distribution of street intersections. This research contrasts previous work, which focused on learning representations using computer vision approaches that utilised convolutions layers and relied on creating raster representations of the street networks before training. Instead, using graph convolutions, we can learn the low dimensional embedding of street networks directly from their adjacency matrix and a feature matrix encoding the spatial distribution of their nodes.

The model learns directly from street network graphs using the graph adjacency matrix and a node attribute matrix as inputs, retaining crucial topological information usually lost when using computer vision models. As a result, the proposed model can generate coherent and diverse street network samples, and the latent representations can be used in downstream tasks such as street network classification.

Despite the positive results, further exploration is necessary. For example, research is still needed to interpret better the learnt latent features and how they relate to the topological and geometric properties of the graphs they encode. Additionally, the model currently is limited to small graphs that only represent a particular place in a city rather than the city as a whole. Since the model requires all graphs to be fixed, entire cities cannot be compared, and the heterogeneity present in urban areas is lost. An immediate implication of the study is that by learning a useful and compact representation from street networks, we can immediately use this information for other downstream geographical tasks, such as prediction or classification. Additionally, by learning a lower-dimensional embedding and the ability to sample from this latent space and generate a synthetic street network, the model can help shed light on the geometric and topological properties of the street networks and complement traditional methods used to study these networks.

## DATA AND CODES AVAILABILITY STATEMENT

The data that support the findings in this study are openly available in OpenStreetMap, and codes that support the findings of this study are available with the identifier(s) at the private link https://figshare.com/s/21dd0d41281e665